\documentclass[runningheads]{llncs}
\usepackage{graphicx}
\usepackage{multirow}
\usepackage{booktabs}
\usepackage{todonotes}
\usepackage{amsmath}
\usepackage{caption}
\usepackage{subcaption}
\usepackage{fontawesome}
\usepackage[backref=page]{hyperref}
\hypersetup{
    pdftitle={velioglu2022xai4pm},
    }

\usepackage{siunitx}
\usepackage{tabulary}

\begin{document}

\title{Explainable Artificial Intelligence for Improved Modeling of Processes\texorpdfstring{\thanks{{\footnotesize This research was supported by the research training group ``Dataninja'' (Trustworthy AI for Seamless Problem Solving: Next Generation Intelligence Joins Robust Data Analysis) funded by the German federal state of North Rhine-Westphalia, and supported by the European Commission Horizon for \textit{ICU4COVID} project, and the VW-Foundation for the project \textit{IMPACT}.}}}.}

\author{
    Riza Velioglu\inst{1,2}\href{https://scholar.google.com/citations?user=bEGGmqgAAAAJ}{(\faGraduationCap)}\
    \and Jan Philip Göpfert\inst{2}\href{https://scholar.google.com/citations?user=rbG7pZQAAAAJ&hl=en}{(\faGraduationCap)}\
    \and André Artelt\inst{1}\href{https://scholar.google.com/citations?user=fbA8-4kAAAAJ}{(\faGraduationCap)}\
    \and Barbara Hammer\inst{1}\href{https://scholar.google.com/citations?user=1d3OxaUAAAAJ&hl=en}{(\faGraduationCap)}\
}

\authorrunning{Velioglu et al.}

\institute{
CITEC, Bielefeld University, Bielefeld, Germany \\ \email{\{rvelioglu,aartelt,bhammer\}@techfak.de} \
\and Recommendy UG, Bielefeld, Germany \\ \email{j.goepfert@recommendy.dev} \
}

\maketitle

\begin{abstract}
In modern business processes, the amount of data collected has increased substantially in recent years.
Because this data can potentially yield valuable insights,
automated knowledge extraction based on process mining
has been proposed, among other techniques, to provide users with intuitive access to the information contained therein.
At present, the majority of technologies aim to reconstruct explicit business process models.
These are directly interpretable but limited concerning the integration of diverse and real-valued information sources.
On the other hand, Machine Learning (ML) benefits from the vast amount of data available and can deal with high-dimensional sources, yet it has rarely been applied to being used in processes.
In this contribution, we evaluate the capability of modern Transformer architectures as well as more classical ML technologies of modeling process regularities, as can be quantitatively evaluated by their prediction capability. 
In addition, we demonstrate the capability of attentional properties and feature relevance determination by highlighting features that are crucial to the processes' predictive abilities.
We demonstrate the efficacy of our approach using five benchmark datasets and show that the ML models are capable of predicting critical outcomes and that the attention mechanisms  or XAI components
offer new insights into the underlying processes.

\end{abstract}

\section{Introduction}
Data is collected on anything, at any time, and in any location. A study by IBM~\cite{ibm2020} found that in 2020, 40~trillion gigabytes (40~zettabytes) were generated.
The majority of data in the digital realm, however, is unstructured, making it impossible for humans to process it in its entirety and leaving businesses struggling to manage such massive amounts of data. As a result, one of today's major issues for businesses is extracting information and value from data contained in information systems.
%

Process Mining~(PM)~\cite{van2016process} is a relatively new area of research that lies between process modeling and analysis, machine learning, and data mining.
It is an umbrella term for a family of techniques to facilitate the analysis of operational processes by extracting knowledge from event logs in symbolic form, typically in the form of process models.
An event log serves as the input to PM algorithms, which is essentially a database of events where each event has
(1) a case id: a unique identifier for a particular process instance,
(2) an activity: description of the event that is occurring, and
(3) a timestamp: timestamp of the activity execution.
Resources, expenses, and other event-related attributes may also be integrated by some techniques.
%
PM techniques mostly address three tasks:
(1) process discovery: transform the event log into a process model which describes the underlying process generating such data: common techniques include the alpha algorithm, heuristic miner, or inductive miner~\cite{van2004},
(2) conformance checking: analyse discrepancies between an event log and an existing process model to detect deviations, for quality checking or risk management~\cite{Munoz},
(3) process enhancement: improve the existing model's performance in terms of specific process performance metrics \cite{enhance}.

The initial focus of PM was the analysis of \textit{historical} event data to generate a process model.
These monitoring systems are reactive in nature, allowing users to detect a violation only after it has occurred. 
In contrast, 
Predictive Process Mining~(PPM)
or Predictive Process Monitoring~\cite{maggi2013}
aims for forward-looking forms of PM with predictive qualities.
It is a field that combines machine learning with process mining, aiming for specific tasks such as
predicting 
\textit{next activity}, \textit{activity suffix}, \textit{next timestamp}, \textit{remaining time}, \textit{attributes}, \textit{attribute suffix}, and \textit{outcome} of running cases.
Most tasks can be modeled as classification problems, except \textit{next timestamp} and \textit{remaining time} prediction tasks, which are regression problems.
Some approaches which have been used in this realm are based on
recurrent neural networks~(RNNs) or Long-short term memory~(LSTMs)
\cite{EVERMANN2017129,tax2017predictive,nguyen2020time}. 
Alternatives are based on 
Autoencoders~\cite{mehdiyev2020novel} and Convolutional neural networks~\cite{pasquadibisceglie2019using}.
More recently, Transformer~\cite{vaswani2017} models have gained a lot attention due to their overwhelming success in computer vision~\cite{dosovitskiy2020image} and natural language processing~\cite{devlin2018bert}.
Inspired by their success, recent approaches proposed a Transformer-based model for process data \cite{bukhsh2021}. 
A more detailed overview on different models for the  aforementioned tasks is covered in the survey~\cite{rama2021deep}.

Previous work in PPM faces a few challenges:
unclear training/test set splits and pre-processing schemes, which leads to noticeably different findings and challenges w.r.t\ reproducibility~\cite{weytjens2021}.
Since datasets often contain duplicates which are not respected by the evaluation schemes,  results are often confounded.
In this work, we investigate the capability of important classical ML technologies as well as modern Transformers on a variety of benchmark sets with clear train/test splits, duplicates respected, and different types of representation for classical schemes.
More importantly, we do not treat the methods
as opaque schemes, but rather rely on feature relevance determination and attention mechanisms to highlight the most important factors for each model.
This is in line with previous approaches such as 
\cite{harl2020explainable}, which 
visualizes the feature relevances of a gated graph neural network 
or
\cite{hsieh2021dice4el} which enhances next activity prediction using LSTMs with counterfactual explanations for wrongly classified samples.
Yet both approaches evaluate the behavior on a single
and specific dataset only.
In our approach, we systematically investigate relevant ML models for different approaches and representations under the umbrella of explainability. In this way, we demonstrate how, in many cases, the presence of
specific events can easily fool the system into relying on `trivial' signals. When these are removed, ML models reveal not only high quality behavior but also intriguing insights into the relevance of more subtle signals or sequences. 
%
\paragraph*{Contributions}
Our contribution is twofold:
a systematic study of the behavior of diverse ML models with 
train-test split respecting the peculiarities of  datasets in PM and appropriate pre-processing,
which prevent troubles due to data leakage.
Second, we utilize eXplainable Artificial Intelligence~(XAI) techniques to compute and visualize the most crucial features.
Our code is open source and available at~\url{https://github.com/rizavelioglu/ml4prom}.

The remainder of this work is structured as follows:
In Section~\ref{methodology}, we present the peculiarities of the datasets and our methodology.
In Section~\ref{experiments}, we present experiments and results.
In Section~\ref{conclusion}, we conclude our contribution and discuss potential directions for future work.

\section{Methodology}\label{methodology}
In this section we describe the datasets and how they are pre-processed to be utilized for binary classification, and highlight a few  domain-specific peculiarities which have to be taken into account.
Then we introduce both classical ML models as well as the state-of-the-art Transformer model used subsequently.
Lastly, we present the XAI techniques applied.

\subsection{Data}
We focus on five widely used datasets which come from a variety of domains including loan applications, road traffic management, and healthcare. These datasets are benchmarks from PM and are also accessible within the well-known ProM tools~\cite{VanDongen2005}, with the exception of healthcare, which contains personal data. Each dataset has been labeled in order to assess explainability and put them in the ML context. The variety of datasets enables us to check the robustness of our methodology across a wide range of domains. To apply ML classifiers to event data, we first transform the data into a format that resembles a binary classification problem. We present five real-life event logs, each with its own evaluation criterion for a prediction task (referred to as the positive/negative subpart of the logs), and thereafter explain how they are transformed for the task.
\paragraph*{BPIC 2017\_application+offer~\cite{vanDongen2017}:} a loan application process of a Dutch financial institute that covers \num{31509}~cases. Each case represents a loan application and every application has an outcome: positive if the offer is accepted and negative if rejected.
\paragraph*{BPIC 2018\_application~\cite{vanDongen2018}:} an agricultural grant-application process of the European Union that contains \num{43809}~applications over a period of three years together with their duration.
\paragraph*{Traffic Fine Management~\cite{deLeoni2015}:} an event log of an information system managing road traffic fines by the Italian government, with fines being payed or not
\paragraph*{COVID~\cite{pegoraro2022analyzing}:} a dataset of COVID-19 patients hospitalized at an intensive care unit
consisting of 216 cases, of which 196 are complete cases~(patients have been released either dead or alive) and 20 ongoing cases~(partial process traces) that were being treated in the COVID unit at the time data was exported. 
\paragraph*{Hospital Billing~\cite{Mannhardt2017}:} an event log of the billing of medical services that have been provided by a regional hospital. Each trace in the event log keeps track of the actions taken to bill a group of medical services. A randomly selected sample of \num{100000}~process instances make up the event log.

\subsection{Encoding}

Table~\ref{table:dataset} shows the rules used to label the samples as well as important statistics. One 
characteristic of typical PM datasets is that 
they can contain a large number of duplicates, \textit{i}.\textit{e}.\ observations of the same process.
For example, \texttt{Hospital} dataset contains \num{41343} and \num{58653} traces in positive and negative classes, respectively. Out of those traces only \num{306} and \num{884} are unique.
In addition, the top-\num{10} most frequent traces make up the \num{95.3}\% and \num{89.7}\% of the whole traces.
Hence such data are challenging for ML due to a narrow variety. In addition, duplicates need to be accounted for in evaluation to avoid information leakage.

\begin{table}
    \centering
    \caption{Datasets statistics and the rules used to generate binary classification task~(L\textsuperscript{+}/L\textsuperscript{-} represent positive/negative logs, respectively).
    The statistics are the number of traces, the number of unique traces, the percentage of unique traces, and lastly the cumulative percentage of the top-10 most frequent traces.}
    \label{table:dataset}
    \footnotesize
\begin{tabular}{c@{\hskip 4mm}l@{\hskip 4mm}l@{\hskip 4mm}S[table-format = 5.0]@{\hskip 4mm}S[table-format = 4.0]@{\hskip 2mm}S[table-format = 2.1]@{\hskip 2mm}S[table-format = 2.1]} \toprule
{Dataset}          & {Class}       & \multicolumn{1}{c}{{Classification Criteria}} & {traces} & {uniq.} & {uniq./\%} & {top-10\%}    \\ 
\toprule
\multirow{2}{*}{BPIC17}   & L\textsuperscript{+} & without activity `A\_incomplete'                     & 16506               & 529    & 3.2    & 83.5    \\
                          & L\textsuperscript{-} & with activity `A\_incomplete'                        & 15003               & 2101   & 14.0   & 51.5    \\
\midrule
\multirow{2}{*}{Traffic}  & L\textsuperscript{+} & end activity `Payment'                               & 67201               & 122    & 0.2    & 99.1    \\
                          & L\textsuperscript{-} & end activity not `Payment'                           & 83169               & 109    & 0.1    & 99.2    \\
\midrule
\multirow{2}{*}{COVID}    & L\textsuperscript{+} & with activity `Discharge alive'                      & 136                  & 33    & 24.0   & 74.3    \\
                          & L\textsuperscript{-} & with activity `Discharge dead'                       & 60                   & 20    & 33.0   & 83.3    \\
\midrule
\multirow{2}{*}{BPIC18}   & L\textsuperscript{+} & duration less than 9 months                          & 27966               & 1081   & 3.9    & 57.7    \\
                          & L\textsuperscript{-} & duration more than 9 months                          & 15843               & 477    & 3.0    & 82.1    \\
\midrule
\multirow{2}{*}{Hospital} & L\textsuperscript{+} & duration less than 3 months                          & 41343               & 306    & 0.7    & 95.3   \\
                          & L\textsuperscript{-} & duration more than 3 months                          & 58653               & 884    & 1.5    & 89.7   \\
\bottomrule
\end{tabular}
\end{table}

\sloppy 
Data consist of sequences of symbolic events of different length. For each of these datasets we compute \(n\)-grams for \(n \in \{1, 2, 3\}\) to encode traces in vectorial form  for classic ML models. Unigrams~(n=1) encode occurrence of events, bigrams~(n=2) encode  two subsequent events and trigrams~(n=3) encode three subsequent events. 
%
Transformer models can directly deal with sequences. The network learns a vector embedding for each event within a trace. 
This has the advantage of avoiding problems caused by the high-dimensionality of one-hot encoding method, for example.

\subsection{Models}
Using these representations, we train a variety of models, including Logistic Regression~(LR), Decision Tree~(DT), Random Forest~(RF), Gradient Boosting~(GB), and Transformer models. We only present the LR, DT and Transformer models because the findings do not vary significantly. We selected LR and DT due to their widespread use and overall high performance for classification problems \cite{fernandezdelgado2014hundred}, and we selected Transformer models due to their propensity for learning complex patterns.
The type of encoding used in this case limits the amount of information that is accessible because only a portion of the sequential structure is represented. Conversely, transformer models can easily handle sequential data. 
We use models as proposed in the works~\cite{vaswani2017,bukhsh2021}.

\subsection{XAI}
Since our primary concern is not the classification performance of our trained models, but rather whether ML models can capture underlying regularities, we employ XAI techniques to gain insight into which relationships between events in the data are used by the  models \cite{molnar2022}.
Unlike many explanation approaches such as LIME~\cite{lime}, LRP~\cite{lrp}, we are interested in global explanations rather than explanations of single decisions. This is because our goal is not to comprehend the ML model, but rather to gain understanding of the key components of the PM as a whole, which can enable users to enhance processes. As an example, one may anticipate altered process behavior if they come across a specific activity, such as `Discharge dead' in COVID dataset, which was identified as crucial for the ML model.
As a result, we strive for global explanations and, more specifically, we employ a number of well-established feature selection methods for both linear and non-linear settings, including LR with LASSO as a linear model with strong mathematical guarantees~\cite{lasso}, DT as a non-linear model with efficient Mean Decrease in Impurity~(MDI) and permutation importance~\cite{understandingrf} that takes into account non-linear relations between features to determine the relevances, and Transformer model equipped with an attention mechanism that highlights complex relationships.
Attention mechanisms are technically local explanations. We utilize them to see if highly flexible non-linear explanations provide more complex relations than global feature importance measures.

As a first result of the analysis, feature selection methods immediately reveal a potential source of information leakage:
Leaving the datasets as they are, 
the algorithms rely on attributes that directly encode the class label but provide no additional insight into the process.
Table~\ref{table:dataset_biased_feats} shows the detected features that are removed from the logs to avoid such trivial outcomes.
All of the listed events exist only in their respective class, \textit{e}.\textit{g}.\ `A\_Incomplete' is present only in L\textsuperscript{-} but not in L\textsuperscript{+}, except `Payment' where the event is present in both classes. Therefore, we removed the event only if it is the last event in a trace from L\textsuperscript{+}, as that is the source of leakage~(see Table~\ref{table:dataset}).

\begin{table}[tb]
    \centering
    \caption{Datasets with features that leak label information.}
    \label{table:dataset_biased_feats}
    \footnotesize
\begin{tabular}{l@{\hskip 4mm}l@{\hskip 4mm}l@{\hskip 4mm}} \toprule
Dataset                 & Class                   & \multicolumn{1}{c}{{Biased feature}} \\
\toprule
BPIC17                  & L\textsuperscript{-}    & `A\_Incomplete'              \\
\midrule
\multirow{2}{*}{Traffic}& L\textsuperscript{-}    & `Send for Credit Collection' \\
                        & L\textsuperscript{+}    & `Payment'            \\
\midrule
\multirow{2}{*}{COVID}  & L\textsuperscript{-}    & `Discharge dead'             \\
                        & L\textsuperscript{+}    & `Discharge alive'            \\
\bottomrule
\end{tabular}
\end{table}

\section{Experimental Results}\label{experiments}
In this section we present the data pre-processing pipeline used to transform all five datasets for training. Then, we present the evaluation metric used as well as the model design and training on each dataset. Finally, we present the scores of different approaches on the binary classification task and the resulting feature relevances. To save space, we limit ourselves to the \texttt{BPIC17} dataset; results for the other datasets can be found in the GitHub repository.

\subsection{Data Split and Pre-Processing}
For simplicity, we only consider event names to encode traces and no other attributes, \textit{e}.\textit{g}.\ timestamp, or resource of an event. After computing unique traces we randomly sample from them to construct train/test sets with a ratio of \num{70}\%/\num{30}\%, respectively. As the datasets are highly imbalanced, we sample class-wise from data--preserving the proportion of classes both in train and test sets. To account for frequency of traces, we keep the duplicate traces in train set while removing the ones in test set. 

We apply minimal pre-processing to data at hand. First, we remove the biased features from traces. Then we add \texttt{<start>} and \texttt{<end>} tokens to the input to explicitly define the beginning and the end of traces. To have a fixed-length input, the traces that are shorter than the longest trace in an event log are padded with the \texttt{<pad>} token. We then build a token dictionary consisting of unique event names in an event log and the aforementioned special tokens. Finally, the tokens in the input are replaced by their unique integer values stored in the dictionary.

\subsection{Evaluation Metric} 
Because the datasets are highly imbalanced, we measure performance using the area under the receiver operating characteristic curve~(AUROC)~\cite{Bradley1997auc}. ROC analysis does not favor models that outperform the majority class while under-performing the minority class, which is desirable when working with imbalanced data~\cite[p.~27]{mehdiyev2020novel}.
The area under the curve of a binary classifier 
is the probability that a classifier would rank a randomly chosen positive instance higher than a randomly chosen negative one, which is given by the following equation:
  \begin{equation}
        \texttt{AUROC} = \int_{x=\infty}^{-\infty} \texttt{TPR}(T) \texttt{FPR}^{'}(T)dT
  \end{equation}
\texttt{AUROC} ranges from \num{0} to \num{1}, with an uninformative classifier~(random classifier) producing a result of \num{0.5}.

\subsection{Model Design and Training}
For ML models we utilize scikit-learn~\cite{pedregosa2011scikit} and initialize the models with default parameters. 
For logistic regression, we employ regularization by  a L1 penalty term(\texttt{penalty} hyper-parameter) with regularization strength~(\texttt{C} hyper-parameter). We use repeated stratified \textit{k}-fold cross-validation~(CV) 
to evaluate and train a model across a number of iterations. 
The number of repeats is \num{50}, and the number of splits $\textit{k}=5$, yielding \num{250} models being trained. The reported metric is then the average of all scores computed.

For Transformer model, following~\cite{bukhsh2021}, we chose the embedding dimension as \num{36}, \textit{i}.\textit{e}.\ each trace is represented by a point in \num{36}-dimensional space. Since Transformer disregards the positional information of events in traces, we add positional encoding to token embedding which have the same dimensions. During training, the model learns to pay attention to input embedding as well as positional encoding in an end-to-end fashion. The embedding outputs are then fed to a multi-head attention block with $h=6$ heads. On the final layer of the attention block, we aggregate features using a global average pooling followed by a dropout at a rate of \num{0.1}. Then, we employ a dense layer with ReLU activation of \num{64} hidden units and a dropout at a rate of \num{0.1}. Finally, we use a dense layer with sigmoid activation that outputs a value between \num{0} and \num{1}, which is interpreted as the ``\textit{probability}'' of a trace belonging to positive~(desirable) class. We train the model for \num{50} epochs with ADAM optimizer~\cite{adam2014}, a learning rate of \num{1e-3}, and batch size of \num{16}.

\subsection{Results}

\subsubsection{Predictive Accuracy}

We report the experimental results for the binary classification task. Table~\ref{table:results} reports the \texttt{AUROC} scores of models on \texttt{BPIC17, Traffic}, and \texttt{COVID} datasets, where there are some features leaking the label information. Here the perfect score is achieved as the models correctly discover the biased features, as expected. However, the Transformer model and the ML models with \num{1}-gram on \texttt{Traffic} dataset do not achieve the perfect score on the test set. This is due to the fact that the biased feature--`Payment'--appears in both of the classes. Therefore, \num{1}-gram encoding is not capable of leaking the information. The Transformer model also fails to discover the bias. On the other hand, models achieve the expected results with the \num{2}-gram and \num{3}-gram encoding because those encoding types explicitly define the biased feature: as \texttt{<end>} token added to traces, the feature \texttt{(Payment, <end>)} in \num{2}-gram and \texttt{(Payment, <end>, <PAD>)} in \num{3}-gram would leak the label information~(positive class if last event is `Payment', negative class otherwise).

\begin{table}
    \centering
    \caption{AUROC scores on datasets \textit{with} and \textit{without} biased features using different encoding methods and models. The values inside the parenthesis represent the score on the hold-out test set, whereas others represent the score on the training set.}
    \label{table:results}
    \footnotesize
    \resizebox{\linewidth}{!}{
\begin{tabular}{ll lll lll} \toprule
                              &                      & \multicolumn{3}{c}{\textbf{with biased features}}                         & \multicolumn{3}{c}{\textbf{without biased features}} \\ \cmidrule(lr){3-5} \cmidrule(lr){6-8}
\textbf{Dataset}              & \textbf{Encoding} \hspace{2mm} & \textbf{LR}      & \textbf{DT} & \textbf{Transformer} \hspace{2mm} & \textbf{LR}       & \textbf{DT} & \textbf{Transformer} \\ 
\toprule
\multirow{4}{*}{BPIC17}       & integer              & -                & -                               & 100.(100.)           & -                 & -                               & 97.3(97.3)          \\
                              & 1-gram               & 100.(100.)       & 100.(100.)                      & -                    & 89.0(81.3)        & 89.5(76.0)                      & -                   \\
                              & 2-gram               & 100.(100.)       & 100.(99.9)                      & -                  & \textbf{97.4(97.9)} & 97.3(93.5)                      & -                   \\
                              & 3-gram               & 100.(100.)       & 99.9(99.8)                      & -                    & 97.4(97.8)        & 97.3(91.7)                      & -                   \\ 
\midrule
\multirow{4}{*}{Traffic}      & integer              & -                & -                               & 100.(98.4)           & -                 & -                               & \textbf{63.9(53.5)}   \\
                              & 1-gram               & 100.(93.3)       & 100.(92.2)                      & -                    & 61.5(36.8)        & 61.5(52.7)                      & -                     \\
                              & 2-gram               & 100.(100.)       & 100.(100.)                      & -                    & 63.8(45.5)        & 63.9(49.8)                      & -                     \\
                              & 3-gram               & 100.(100.)       & 100.(100.)                      & -                    & 63.8(48.1)        & 63.9(52.3)                      & -                     \\ 
\midrule
\multirow{4}{*}{COVID}        & integer              & -                & -                               & 90.9(100.)           & -                 & -                               & \textbf{74.8(94.2)}   \\
                              & 1-gram               & 100.(100.)       & 100.(100.)                      & -                    & 67.3(75.0)        & 69.6(68.3)                      & -                      \\
                              & 2-gram               & 100.(100.)       & 99.9(100.)                      & -                    & 89.3(61.7)        & 85.4(85.0)                      & -                      \\
                              & 3-gram               & 99.9(96.7)       & 98.1(66.7)                      & -                    & 90.4(48.3)        & 85.5(76.7)                      & -                      \\ 
\midrule
\multirow{4}{*}{BPIC18}       & integer              & -                & -                               & -                    & -                 & -                               & 98.6(81.2)            \\
                              & 1-gram               & -                & -                               & -                    & 97.5(77.9)        & 97.5(78.5)                      & -                       \\
                              & 2-gram               & -                & -                               & -                  & \textbf{98.4(87.3)} & 98.2(84.5)                      & -                       \\
                              & 3-gram               & -                & -                               & -                  & \textbf{98.4(87.3)} & 98.2(81.5)                      & -                        \\ 
\midrule
\multirow{4}{*}{Hospital}     & integer              & -                & -                               & -                    & -                 & -                               & \textbf{92.4(78.7)}     \\
                              & 1-gram               & -                & -                               & -                    & 91.8(73.9)        & 92.3(56.2)                      & -                        \\
                              & 2-gram               & -                & -                               & -                    & 92.5(70.5)        & 92.7(52.5)                      & -                        \\
                              & 3-gram               & -                & -                               & -                    & 92.6(66.0)        & 92.6(48.5)                      & -                        \\
\bottomrule
\end{tabular}}
\end{table}

Table~\ref{table:results} also reports the \texttt{AUROC} scores of models on all of the five real-life event logs, where the biased features are removed from \texttt{BPIC17, Traffic}, and \texttt{COVID} datasets. For \texttt{BPIC17} and \texttt{COVID} datasets the scores worsen but the results are still promising, hinting that the processes, sequences of events, maintain valuable information for the task even though the distinct/biased features are removed. 
The different amount of information in different encoding is mirrored by the results: \num{1}-gram achieves the worst results compared to \num{2}-gram and \num{3}-gram, as it only incorporates single events, whereas \num{2}-gram and \num{3}-gram incorporates event pairs. None of those methods integrate order information, while Transformer model learns this information during training. We observe that Transformer model successfully captures the relations between events as well as the order information and outperforms other models in \texttt{Traffic} and \texttt{COVID} datasets, whereas in \texttt{BPIC17} it receives a comparable result. On the other hand, we observe that the scores for \texttt{BPIC17} and \texttt{COVID} datasets do not fluctuate as much as it does for \texttt{Traffic} dataset when compared to biased scores represented in Table~\ref{table:results}. This is due to the fact that after removing the biased feature in \texttt{Traffic} dataset, some traces in both classes become identical. 



\subsubsection{Feature Relevances}

\begin{figure}
    \centering
    \includegraphics[width=.55\textwidth]{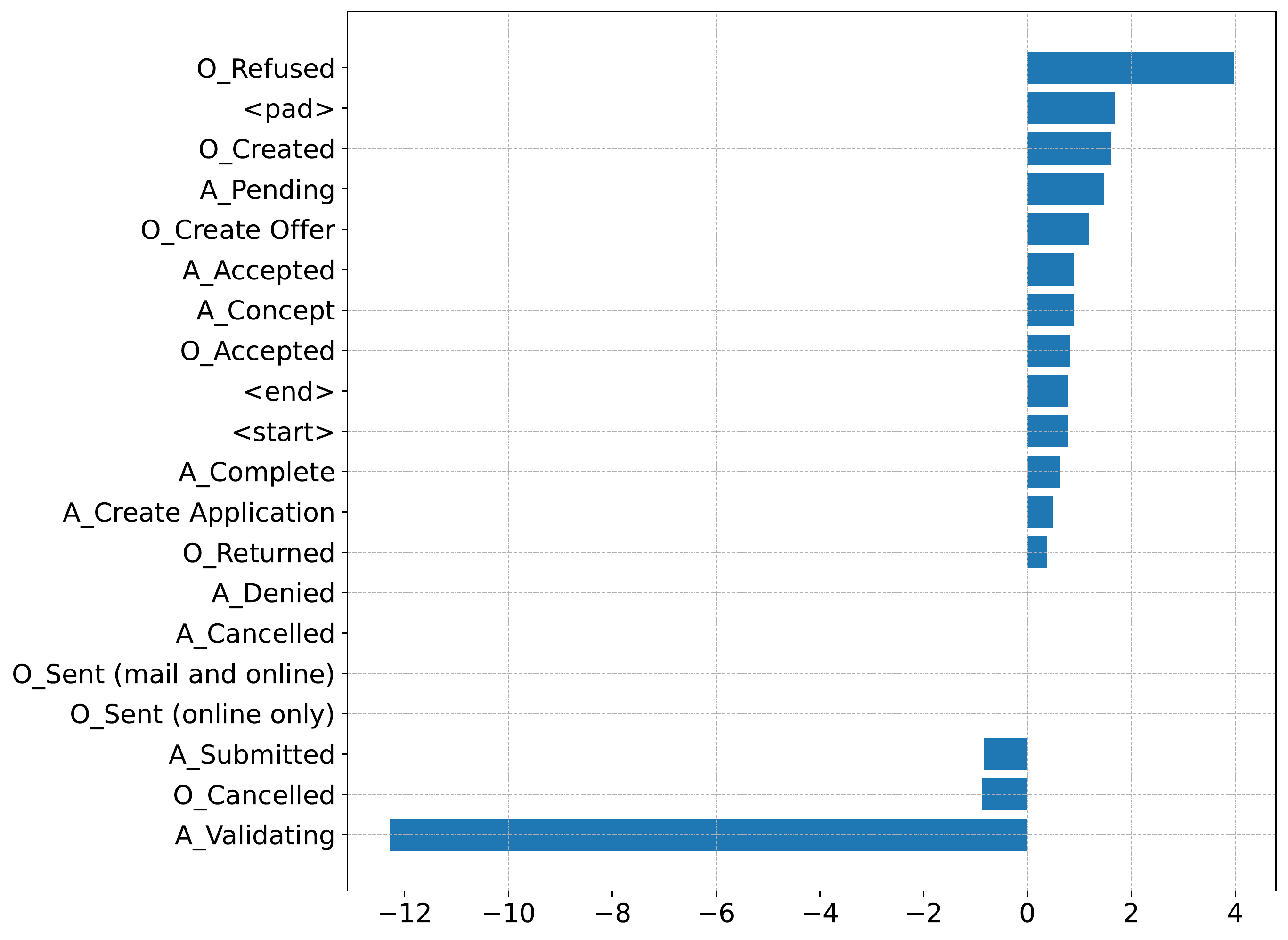}
    \vspace{-2mm}
    \caption{Relevance of features as considered by the logistic regression model. 
    A high positive/negative value indicates a considerable contribution towards predicting the positive/negative label, \textit{i}.\textit{e}. desirable/undesirable event trace.}
    \label{fig:relevances-unbiased-logistic-regression}
\end{figure}

We present the relevances of features taken into account by our trained logistic regression model in Figure~\ref{fig:relevances-unbiased-logistic-regression}. We find that `A\_Validating' contributes the most towards predicting an undesirable trace, whereas `O\_Refused' contributes most to predict a desirable trace. Other events have significantly lower impact on the predictions. Interestingly, the special tokens, \textit{i}.\textit{e}. \texttt{<pad>}, \texttt{<start>}, and \texttt{<end>} affect the predictions when they should not, despite the fact that their influence is negligible. In addition, some features have no effect on the predictions, \textit{e}.\textit{g}.\ `A\_Denied', `A\_Cancelled'.

\begin{figure}
    \centering
    \includegraphics[width=\textwidth]{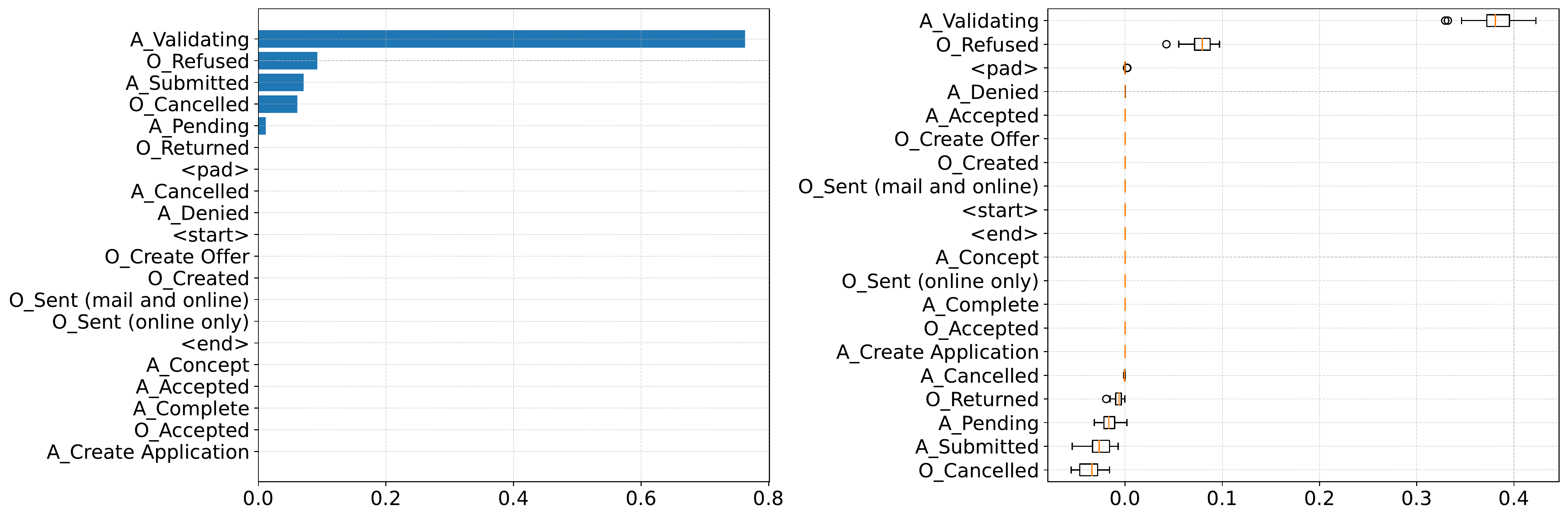}
    \vspace{-6mm}
    \caption{Relevance of features as considered by decision tree model.
    \textbf{Left}: Relevances according to MDI:
    A high value indicates a considerable contribution towards deciding between the positive and negative label.
    \textbf{Right}: Relevances according to the permutation importance calculated on the test set, \textit{i}.\textit{e}.
    how much shuffling a given feature negatively impacts the performance of the decision tree.
    }
    \label{fig:relevances-unbiased-decision-tree}
\end{figure}

In Figure~\ref{fig:relevances-unbiased-decision-tree} we show the feature relevances that our trained decision tree model considers. Based on the MDI value~(left figure) the most relevant feature is `A\_Validating', followed by `O\_Refused', `A\_Submitted', and `O\_Cancelled' with significantly lower impact. Based on the permutation feature importance~(right figure) the most relevant features are `A\_Validating', and `O\_Refused', where the rest of the features have no effect on the prediction performance, which aligns well with the relevances of LR model.

\begin{figure}
     \centering
     \begin{subfigure}[b]{0.4\textwidth}
         \centering
         \includegraphics[width=\textwidth]{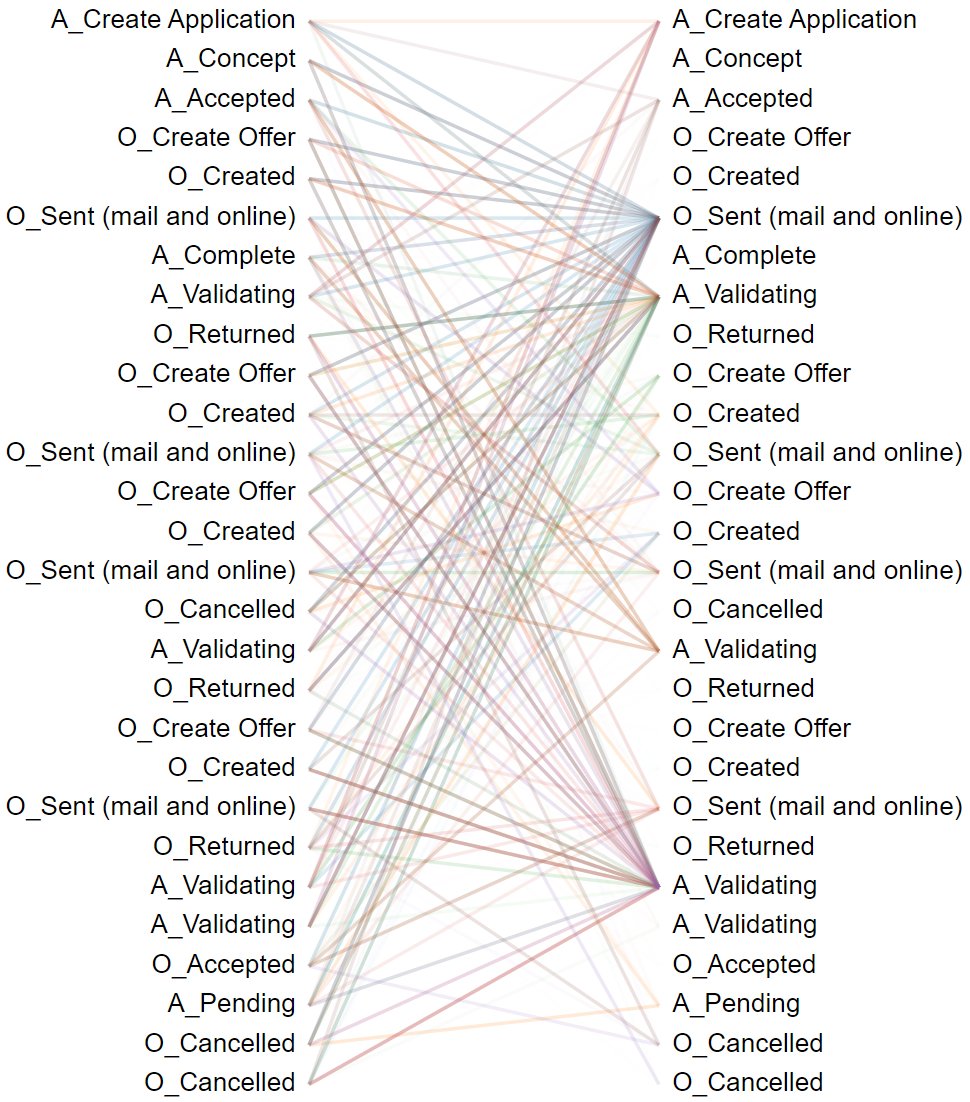}
         \label{fig:y equals x}
     \end{subfigure}
     \hfill
     \begin{subfigure}[b]{0.5\textwidth}
         \centering
         \includegraphics[width=\textwidth]{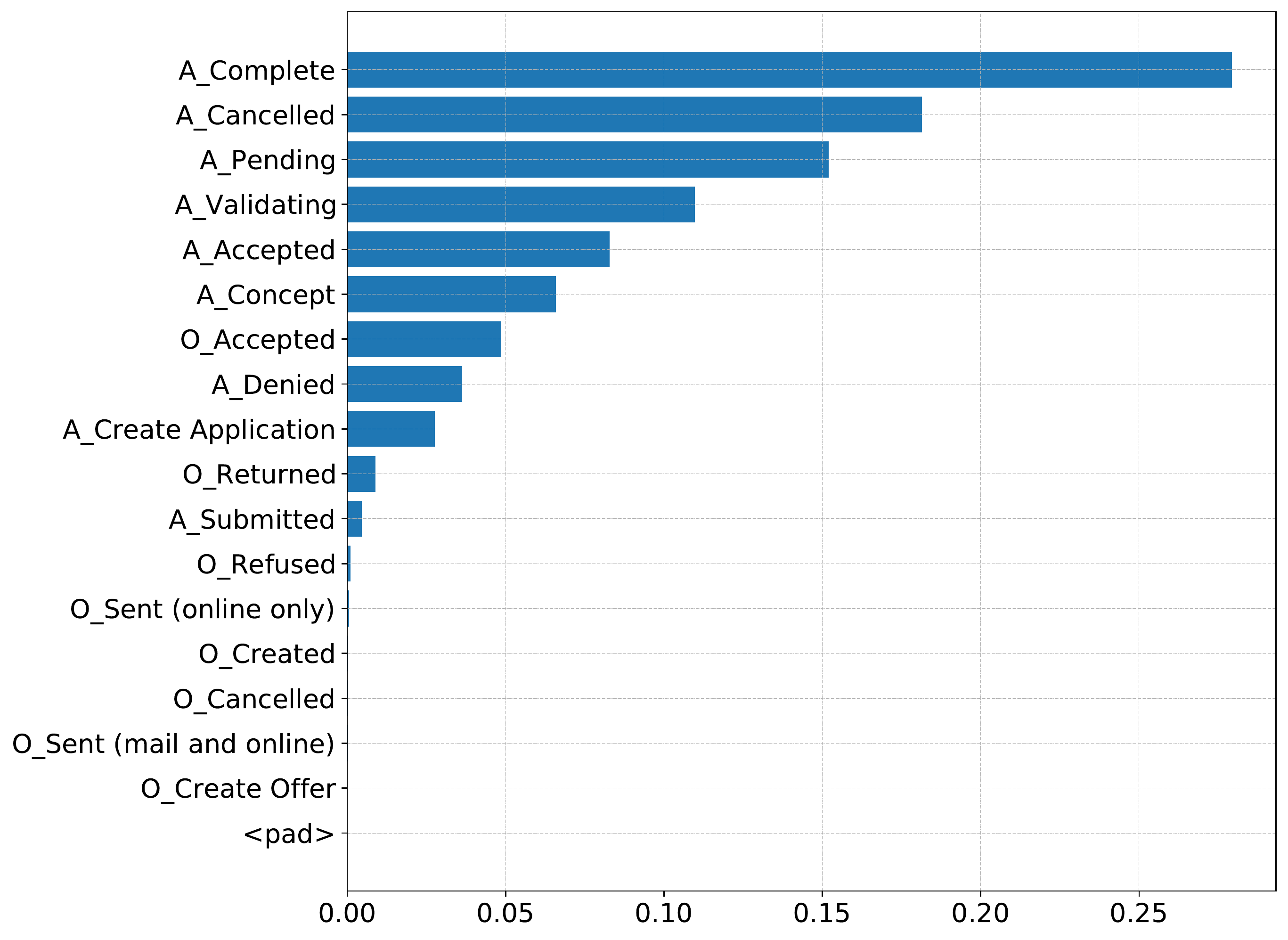}
         \label{fig:three sin x}
     \end{subfigure}
     \vspace{-5mm}
        \caption{
        \textbf{Left}: Relevance of features as considered by transformer model for one randomly selected event trace. The thickness of a line connecting two features indicates the intensity of attention in between, whereas the color represents one of the six different attention heads. 
        \textbf{Right}: Normalized attention scores which are averaged among all attention heads and all traces in test set.}
        \label{fig:relevances-unbiased-transformer}
\end{figure}

Figure~\ref{fig:relevances-unbiased-transformer} visualizes the attention scores of six attention heads for a given trace, as well as normalized attention scores over all attention heads and test samples. We observe that, regardless of where it appears in the trace, the events mostly attend to `A\_Validating' event for the aforementioned trace. In addition, some events focus on `O\_Sent' in some heads, which differs from the other models. The normalized attention scores, however, demonstrates that not all traces exhibit this behavior. The plot also highlights features whose importances overlap with other models' results. In summary, all models agree on the most relevant features.


\section{Conclusion}\label{conclusion}
We have demonstrated how to prepare process data in such a way that it can be used to train classic and modern -- state-of-the-art -- ML classifiers.
All our trained models exhibit high classification performance, \textit{i}.\textit{e}.\ they are capable of learning the underlying regularity of the observed processes, whereby Transformers benefit from the fact that the full sequence information is available, unlike \textit{e}.\textit{g}.\ 1-gram representations. XAI technologies prevent pitfalls such as information leakage by explicit encoding of the predicted event, and reveal insights into the relevance of events or sequences of events, respectively.
These insights enable a further exploration of crucial aspects of the processes, which is useful \textit{e}.\textit{g}.\ for the improvement or correction of undesired process outcomes.

Future research could investigate the effects of various model architectures and encoding schemes on the outcomes of feature relevances. Another potential research direction might be to study the impact of adding further event attributes on the learnt representations.

\bibliographystyle{splncs04}
\bibliography{bibliography}

\end{document}